\documentclass[11pt,a4paper]{article}
\usepackage{graphicx}
\graphicspath{ {images/} }
\usepackage{tabularx}
\usepackage{pgfplots}
\pgfplotsset{width=7.5cm,compat=1.9}
\usepackage{CJKutf8}
\usepackage[hyperref]{acl2020}
\usepackage{times}
\usepackage{latexsym}

\usepackage{microtype}

\aclfinalcopy 

\title{Bil-DOS: A Bi-lingual Dialogue Ordering System (for Subway)}

\author{Zirong Chen \\
  Georgetown University  \\
  Department of Computer Science \\
  \texttt{zc157@georgetown.edu} \\\And
  Haotian Xue \\
  Georgetown University  \\
  Department of Computer Science \\
  \texttt{hx82@georgetown.edu} \\}

\date{}

\begin{document}
\maketitle
\begin{abstract}
Due to the unfamiliarity to particular words(or proper nouns) for ingredients, non-native English speakers can be extremely confused about the ordering process in restaurants like Subway. Thus, We developed a dialogue system, which supports Chinese(Mandarin)\footnote{In this paper, term \emph{Chinese} refers to \emph{Chinese(Mandarin)}.} and English\footnote{All codes are available online at \href{https://github.com/georgetown-dialogue-systems-2020/Bi-Lingual-Dialogue-System}{our GitHub Repo}.} at the same time. In other words, users can switch arbitrarily between Chinese(Mandarin) and English as the conversation is being conducted. This system is specifically designed for Subway ordering\footnote{See menu at \href{https://www.subway.com/en-US/MenuNutrition/Menu/BreadsAndToppings}{Subway website}.}. In Bil-DOS, we designed a {\small\verb|Discriminator|} module to tell the language is being used in inputted user utterance, a {\small\verb|Translator|} module to translate used language into English if it is not English, and a {\small\verb|Dialogue Manager|} module to detect the intention within inputted user utterances, handle 'outlier' inputs by throwing clarification requests, map detected \emph{Intention} and detailed \emph{Keyword}\footnote{In our project settings, we assume there have to be an \emph{Intention} and a \emph{Keyword} within every user utterance, like 'Bread' as \emph{Intention} and '9-Grain Wheat' as \emph{Keyword}.} into a particular intention class, locate the current ordering process, continue to give queries to finish the order,  conclude the order details once the order is completed, activate the evaluation process when the conversation is done.
\end{abstract}

\section{Introduction}
From personal previous experiences, non-native English speakers can find Subway ordering greatly difficult at the first time due to the unfamiliarity to particular words(or proper nouns) for ingredients. Although most of the system we encountered and implemented so far are in English\footnote{Some of them support other languages as well, but we did not have the chance to develop our own version}, we still would like to take a step further to develop a bi-lingual dialogue system, in order to make events like ordering more convenient for non-native English speakers.

In  Bil-DOS,  users  can switch between two languages as the conversation goes on. And at the same time, the system will be able to react to each user utterance using the same language that was most recently used by the user. And after the conversation ends, the system will report order details. Here is happy case\footnote{Unhappy cases of user utterances include: typo, unseen \emph{Intention}s, unseen \emph{Keyword}s.} that the system is supposed to handle:

\begin{figure*}
    \includegraphics[width=0.99\textwidth]{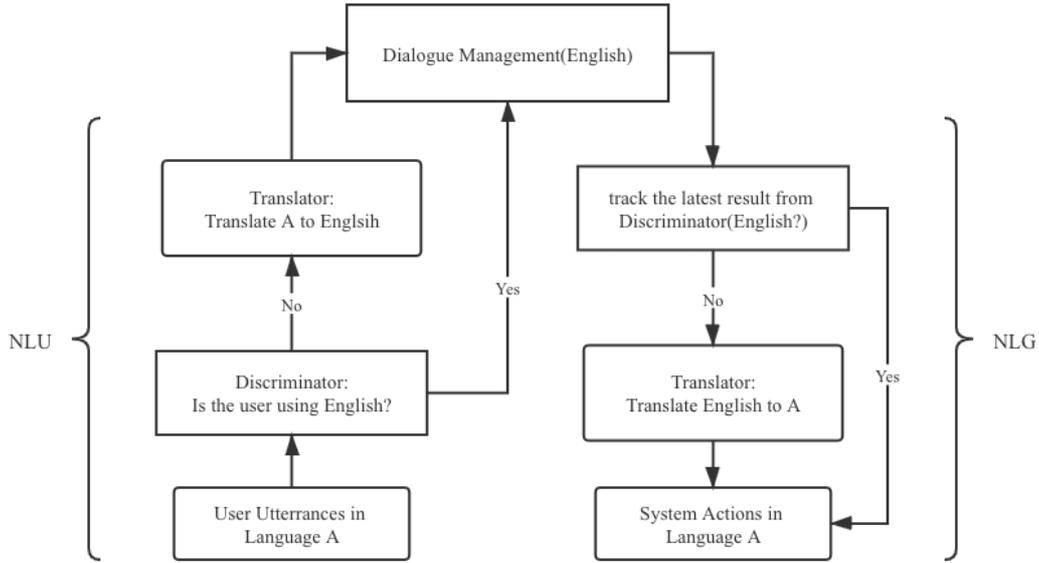}
    \centering
    \caption{Overall structure of Bi-Lingual Dialogue System}
    \label{fig:Overall Structure}
\end{figure*}

[1]  \emph{\textbf{User}}: Hi there!

[2] \emph{\textbf{System}}: Hi, welcome to our Bi-Lingual Ordering System! What can I do for you? Any bread you prefer?

[3]  \emph{\textbf{User}}: Italian bread please!

[4] \emph{\textbf{System}}: Nice choice, you ordered Italian bread as bread, anything else for cheese? 
\begin{CJK*}{UTF8}{gbsn}

[5]  \emph{\textbf{User}}: 羊奶奶酪。(Feta cheese.)

[6] \emph{\textbf{System}}: 您刚刚点了羊奶奶酪作为奶酪，还有什么想要的蔬菜吗？(You just ordered feta cheese as cheese, what do you want for vegetable?)

[7]  \emph{\textbf{User}}: 牛油果。(Avocado.)

[8] \emph{\textbf{System}}: 您刚刚点了牛油果作为蔬菜，还有喜欢的的酱料吗？(You just ordered avocado as vegetable, any preferred sauce?)
\end{CJK*}

[9]  \emph{\textbf{User}}: Barbecue sauce.

[10] \emph{\textbf{System}}: Nice choice, you just ordered barbecue source as sauce, is there any extra you want?

[11]  \emph{\textbf{User}}: No, thanks!

[12] \emph{\textbf{System}}: Okay, you just ordered nothing for extra. I think that is all for your order!

[13] \emph{\textbf{System}}: Fantastic, your order is: one Italian bread sandwich with feta cheese as cheese, avocado as vegetable, barbecue sauce as sauce and with extra Nothing! Thanks for visiting!

....\emph{Evaluation Process Begins Afterwards}...

\section{Related Work}
We model Bil-DOS as a task-oriented dialogue system \citep{gravano2011affirmative, jurafsky2000speech, gao2018neural}. Bil-DOS follows the generally-used \emph{Natural Language Understanding}(NLU)-\emph{Dialogue Management}(DM)-\emph{Natural Language Generation}(NLG) structure. Furthermore, a {\verb|Discriminator|} module, a {\verb|Translator|} module and a {\verb|Dialogue Manager|} module are employed to coordinate with each other and to better execute the `mission' given by the user. The overall structure of our dialogue system is shown in Figure 1.

\subsection{Natural Language Understanding}
An \emph{Natural Language Understanding}(NLU) module is usually built to let the {\verb|Dialogue Manager|} better `understand' human language. In our case, the {\verb|Dialogue Manager|} module is designed and implemented for English only, so, as shown in the figure 1, our NLU module integrates a {\verb|Discriminator|} module to discriminate the language being used and a {\verb|Translator|} module to translate non-English language to English. Only by doing this, can the {\verb|Dialogue Manager|} understand non-English language, which is Chinese(Mandarin) in this case, and handle user requests in both languages.

\subsection{Dialogue Management}
Inspired by Jurafsky's \citep{jurafsky2000speech} and Gao's \citep{gao2018neural} book, our Dialogue Management(DM) module is drafted and implemented in a slot-filling fashion\footnote{In Bil-DOS, the slot is \emph{Intention} and the slot value is \emph{Keyword}.}. Typically, for Subway ordering, there are \emph{slots}, like bread, cheese and sauce, need to be filled as the conversation goes on. By DM \emph{request}ing slot values and the user \emph{inform}ing slot values, the order details will be further supplemented. Once the order details are fulfilled, the conversation session will be terminated and a conclusion will be drawn.

\subsection{Natural Language Generation}
The generation of human language from computer language can be viewed as a general definition of NLG. And to simplify the task, inspired by AIML \citep{bush2001artificial, abushawar2015alice}, Bil-DOS applies pre-written templates to emit actions, in other words, Bil-DOS is a template-based dialogue system.

\subsection{Case Analysis}
Take turn[2] and turn[3] in the conversation sample as the case here. The detailed co-operations between NLU, DM and NLG modules are shown in Figure 2. 

\begin{figure*}[t]
    \includegraphics[width=0.8\textwidth]{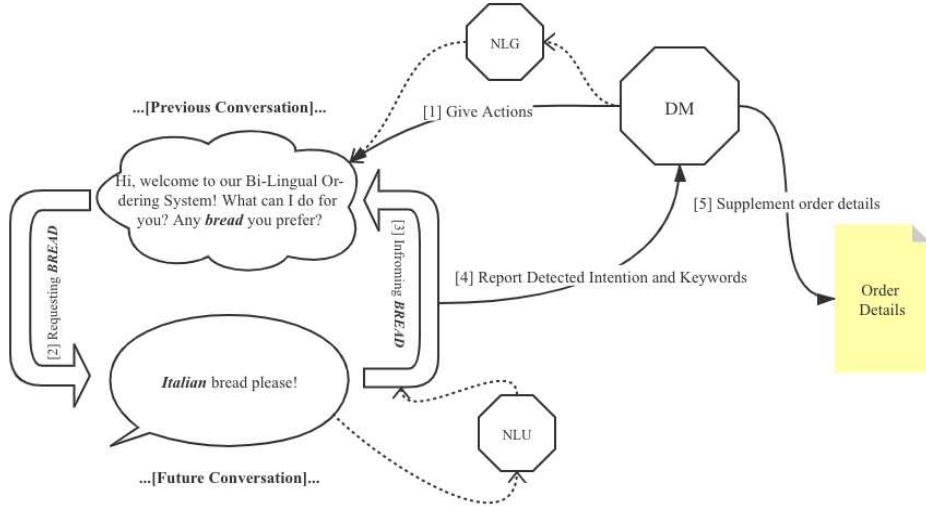}
    \centering
    \caption{Co-operations between modules in Bil-DOS}
    \label{fig:Co-operations}
\end{figure*}

\section{Approach and System Design}
To realize the the \emph{Natural Language Understanding}(NLU)-\emph{Dialogue Management}(DM)-\emph{Natural Language Generation}(NLG) structure, Bil-DOS employs a {\verb|Discriminator|} module, a {\verb|Translator|} module and a {\verb|Dialogue Manager|} module. In this section, the detailed design of each module are shown in the UML class diagram in Figure 3 and will be discussed. 

\begin{figure*}[t]
    \includegraphics[width=0.99\textwidth]{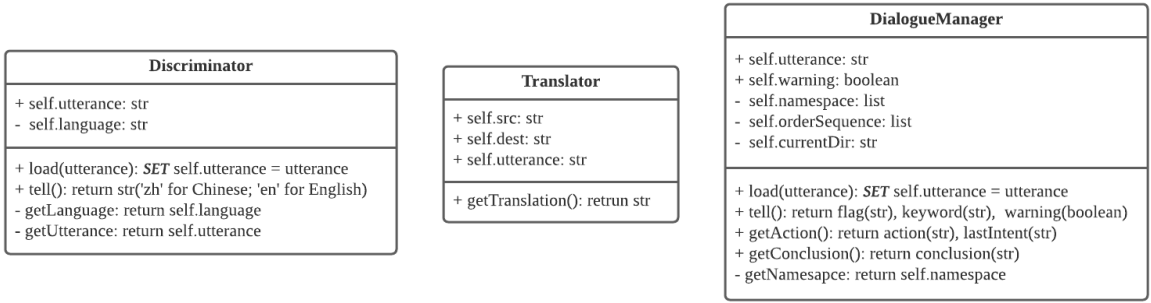}
    \centering
    \caption{UML class diagram for modules}
    \label{fig:UML class diagram}
\end{figure*}

\subsection{The Discriminator}
This module is designed for turn-level language detection. Current methods include rule based detection \citep{buryak2015rules}, Bayes Theorem approximation \citep{jimenez2006text}, cluster-based inference \citep{buryak2015cluster}, neural networks \citep{kim2014temporal}. However, most of these methods are dealing with languages that at least share some part of their alphabets. In our case, English and Chinese have totally different alphabets, then character-level Unicode \citep{allen2012unicode} comparison\footnote{In Unicode comparison, once a Chinese character shows up, the user utterance will be marked as '\emph{zh}'.} will be sufficient to tell the differences.

The {\verb|Discriminator|} module has {\small\verb|language|}, {\small\verb|utterance|} as its attributes\footnote{N.B. All the attributes are predefined with the class initialization function}, and {\small\verb|load(.)|}, {\small\verb|tell()|} as its public functions.

The {\small\verb|language|} attribute tells the supported languages and currently it is `\emph{en\_zh}' by default, which means the {\verb|Discriminator|} supports English('en') and Chinese('zh'). The {\small\verb|utterance|} attribute is used to store current user utterance and will be set to \emph{None} upon another time of use. 

The {\small\verb|load(.)|} function takes user utterance once a time and store it for further use, once one turn is over, the storage will be evicted. The {\small\verb|tell()|} is able to tell what the language is being used from the most recent user utterance.  

\subsection{The Translator}
The {\verb|Translator|} module is wielded to translate non-English language, which is Chinese(Mandarin) in our case, to English. We first explored the possibility to apply BERT \citep{devlin2018bert} language models using Hugging Face package \citep{wolf-etal-2020-transformers}. However, after several attempts to implement a pipeline for language translation, this method turned out to be far more time-consuming than we expected. For our task, a more efficient approach is needed for conversation fluency\footnote{In the process, we do not want any palpable lagging between the order system and the user.}. Thus, another solution for translation tasks is found: (packages or API of)Online translators. The package used is named as \emph{translators} \citep{UlionTse2020}. This package is a free, online and toy-level library that helps users to send translation requests to some famous online translators, like Google, Yandex, Microsoft(Bing)and Baidu, and then fetch the translation results from those translators. The {\verb|Translator|} module in Bil-DOS is built on top of the \emph{translators} package\footnote{After experiments, this package fails to handle extremely frequent requests and there exists a time limit(approximately 10 minutes) for each session. As stated by the author, this package is only for individual use. For more stable translation service, please refer to official APIs.} .

The {\verb|Translator|} module has {\small\verb|src|}, {\small\verb|dest|}, {\small\verb|utterance|} as its attributes, and {\small\verb|getTranslation()|}as its public function.

The {\small\verb|src|} and {\small\verb|dest|} attributes are used to describe the \emph{source} and \emph{destination} languages that need to be clarified in the translation process. And again, the translation process is turn-level, thus it needs {\small\verb|utterance|} attribute to store the user utterance for further translation. 

The {\small\verb|getTranslation()|} function returns the translated English sentence(s) from the store user utterance if the utterance is not in English. 

\subsection{The Dialogue Manager}

The Dialogue Management module serves as the `heart' of a dialogue system. In Bil-DOS, the DM takes user utterances as input and gives system actions according to the user requests as output. Furthermore, Bil-DOS's Dialogue Manager is able to detect user intentions, map and extract keywords, fulfill order details, and handle error inputs.

The {\verb|Dialogue Manager|} module wields {\small\verb|utterance|}, {\small\verb|namespace|}, {\small\verb|orderSequence|}, {\small\verb|currentDir|}, {\small\verb|warning|} as its attributes, and {\small\verb|tell()|}, {\small\verb|getAction()|}, {\small\verb|getConclusion()|} as its public functions.

The {\small\verb|currentDir|} is directory where the order details are placed and {\small\verb|namespace|} is a list of the names of all the currently stored user intentions stored in the directory `\emph{/IntentDetails}'. The {\small\verb|orderSequence|} is the default ordering logic, which is `\emph{bread}', `\emph{cheese}', `\emph{vegetable}', `\emph{sauce}' and `\emph{extra}'. Although is is named as sequence, user can still break the order sequence as he/she wants in some cases\footnote{The sequence was broken into pairs of two intention, as long as the intention is not `\emph{extra}', the conversation will not end and will start at the intention just entered.}. The {\small\verb|warning|} is a boolean variable to indicate whether there are any `error' inputs detected. And again, the dialogue management process is turn-level, thus it needs {\small\verb|utterance|} attribute to store the user utterance for further analysis. 

The {\small\verb|load(.)|} serves for the same purpose as the one in the {\verb|Discriminator|} module. The {\small\verb|tell()|} function returns the matched intention in the user utterance, the keyword detected under that intention and a boolean value to indicate whether there is a unseen intention involved. The {\small\verb|getAction()|} returns system actions and the last system request. The {\small\verb|getConclusion()|} reads the order details stored under {\small\verb|currentDir|} and returns the order summary.

 \textbf{Intention and Keywords Detection} in Bil-DOS is to detect the intention within the user utterance, in order to interpret what the user is trying to inform by mapping and extracting keywords. There are lots of methods to detect intention or keywords, like neural networks \citep{hussein2002intention}, TF-iDF based evaluations \citep{rose2010automatic}, SVM based methods \citep{zhang2006keyword} or Attention Mechanism \citep{vaswani2017attention}, in Bil-DOS, the intention and keywords detection needs to be much easier to ensure the conversation fluency. Since Bil-DOS is designed for ordering, a menu should be presented at the same when Bil-DOS is being used. Thus, for this kind of consideration, \emph{intention} and its list of \emph{keyword}s are pre-loaded manually into the directory `\emph{/IntentDetails}' according to the information on the menu, like `\emph{bread}' and [`\emph{Italian}', `\emph{9-Grain Wheat}', etc]. As long as the user utterance is loaded by the {\small\verb|load(.)|} function, the {\small\verb|tell()|} function is going to turn the utterance to lower cases to increase the match rate, and iterate through all the \emph{intention}s in the {\small\verb|namespace|}, then read the \emph{keyword}s stored under its specific files, and return the matched \emph{intention} and \emph{keyword}.
 
  \textbf{None and `Invalid'\footnote{The word `invalid' is not telling the inputs themselves are unacceptable, we use this word for the conclusion of all unseen and undetectable inputs for Bil-DOS so far.} Inputs} are very common and difficult issues for template-based dialogue systems. The ability to handle all kinds of inputs is required to boost the user experiences. In Bil-DOS, the conservation is assumed in a particular order, which is `\emph{bread}', `\emph{cheese}', `\emph{vegetable}', `\emph{sauce}' and `\emph{extra}', and that means, for \textbf{None-type inputs}, if a decline intention, which can be '\emph{No}' or `\emph{Nope}' is encountered, Bil-DOS needs to backtrack the last requested slot in the last system action and then fill the word `\emph{Nothing}' into it. And that is why the {\small\verb|getAction()|} function needs to return the intention in the last system request. For \textbf{`invalid' inputs}, Bil-DOS will start a human-in-the-loop annotation process to ask the user to annotate the inputted utterance. And the generated data will be stored in the `\emph{/IntentDetails}' directory for further use, so theoretically, if there is not mis-labelling or too many duplicates, Bil-DOS will be more and more robust to all kinds of user utterance as the time of use increases. The robustness of Bil-DOS is shown in Figure 4 and 5. First, let's try not to be confused about what the `Japanese bread' really is. As shown in Figure 4, the currently stored intentions and keywords failed to handle the input `Japanese bread', so Bil-DOS threw two annotation questions and the annotation results were stored in `\emph{/IntentDetails/bread.txt}' for further use. In Figure 5, for another time, the `Japanese bread' showed up. However, since there were relevant annotations about `Japanese bread' in previous uses, Bil-DOS was able to detect `Japanese bread' this time. In conclusion, Bil-DOS is able to take \textbf{ALL} kinds of user utterances, as long as the user wants to annotate, the intention directory is well maintained, and the conversation is allowed to continue.
  
\begin{figure*}[t]
    \includegraphics[width=0.99\textwidth]{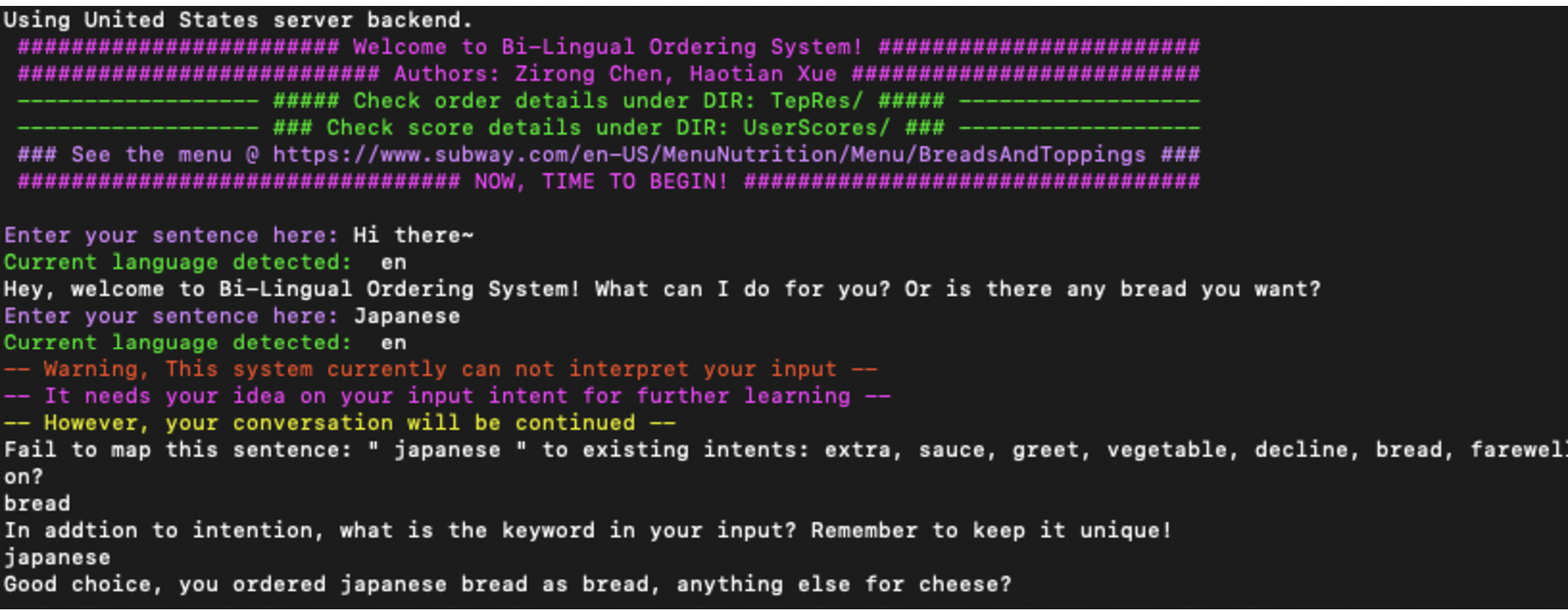}
    \centering
    \caption{The unhappy case in Bil-DOS}
    \label{fig:Unhappy case}
\end{figure*}

\begin{figure*}
    \includegraphics[width=0.8\textwidth]{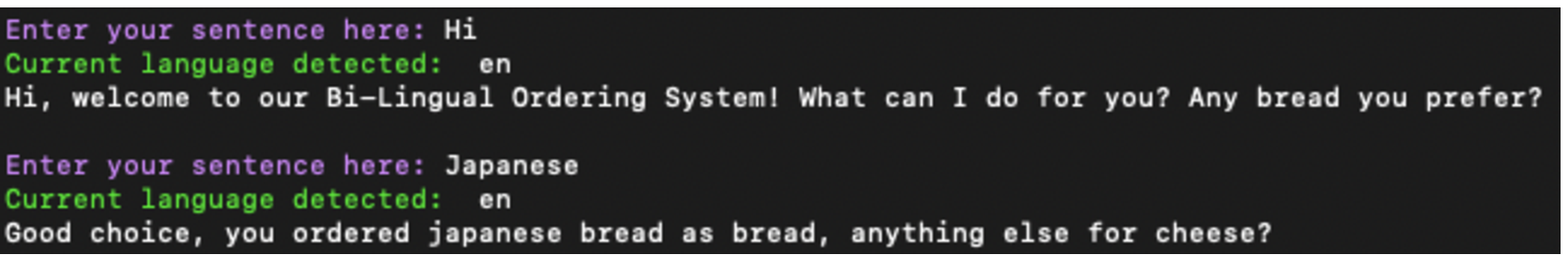}
    \centering
    \caption{The same unhappy case after `learning'}
    \label{fig:Learned unhappy case}
\end{figure*}

\begin{figure*}
    \includegraphics[width=0.8\textwidth]{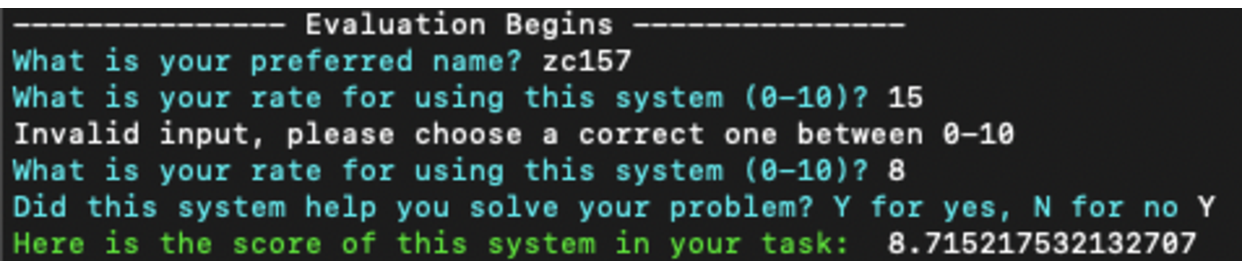}
    \centering
    \caption{Evaluation process in Bil-DOS}
    \label{fig:Evaluation}
\end{figure*}

\subsection{The Evaluator}
The {\verb|Evaluator|} module is built for the evaluation after each conversation. It supports multiple users and stores user scores for each user in his/her file locally. The evaluation process in Bil-DOS is inspired by Gao's book \citep{gao2018neural}. In general, we need the evaluation process to capture the system performance in solving user tasks with least resource, which means we need Bil-DOS to complete the ordering process with as few turns as possible, or in other words, Bil-DOS needs to make every turn most helpful for the task completion. So the key standards are `how well the system helped the user to complete the task' and 'how many turns the system consumed'. Thus, in Bil-DOS built-in evolution module, we prepared parameters like \emph{num\_of\_turns}, \emph{task\_reward}, \emph{turn\_penalty}, \emph{user\_experience} and \emph{score\_factor}. The \emph{task\_reward} is a fixed value, 20 by default, by the end of each conversation, if the user's goal is reached, which is to order a sandwich, then the system will get +\emph{task\_reward}, otherwise it will get -\emph{task\_reward} as penalty. The \emph{turn\_penalty} is set to -1 by default. Once a turn is consumed, the system will get one point off as penalty. The intuition behind is to make the conversation shorter and more efficient. The \emph{user\_experience} is for user subjective evaluation with the scale from 0 to 10. The \emph{score\_factor} is used to balance the weights between user subjective evaluation and Bil-DOS built-in self evaluation, and Sigmoid is used to smooth the results. The reason why Sigmoid function was applied here is that we want both scores taken into consideration without losing the balance. Sigmoid squeezes all numbers between 0 and 1, and it is able to handle extremely large \emph{score\_factor} without losing the consideration of the score from built-in self evaluation. The real evaluation process is shown in Figure 6. The evaluation function\footnote{\emph{Sc} stands for Score, \emph{uSc} stands for user\_score, \emph{scFac} stands for score\_factor, \emph{tuSc} stands for turn\_score, \emph{Sc}taSc stands for task\_score.} is:

\[ Sc = uSc \times scFac + (tuSc + taSc) \times (1 - scFac)\]

\subsection{The Main Session}
The main session for Bil-DOS is similar to Figure 1. However, there are several crucial parameters for conversation restraints and future evaluation. First, \emph{num\_of\_turns}. Since Bil-DOS is a task-oriented dialogue system, there needs to be a specific restraint of the number of turns. By doing so, Bil-DOS will terminate the conversation if the \emph{num\_of\_turns} is consumed. Second, \emph{task\_reward}, \emph{turn\_penalty} and \emph{score\_factor} need to be clarified for the Evaluator module. The last one is \emph{translator}, which is designed in the latest beta version(has not been released yet) and is able to give user options to select different translator backends. The detailed main session is shown in the UML sequence diagram\footnote{The diagram is not a rigorous UML sequence diagram. To better illustrate the cooperation between modules as a sequence, the main session was also placed into this diagram.} in Figure 7.

\subsection{User Interfaces}
Bil-DOS has a poly-chromatic \citep{joeld2008terminal} UI by adding additional statements in Python3 terminal outputs. Because we think a poly-chromatic can grab more attention from the users, express the `feelings' behind system messages, like blue as welcoming, green as ok-ish, red as warning, ect, and even make the system more vivid. The experiment results do show the boost of user scores after applying poly-chromatic UI.

\section{Experiments and Improvements}
In this section, we organized several controlled variable experiments about different system designs. All the participants are Chinese speakers and non-native English speakers, and they are target users that Bil-DOS aims to serve.

\subsection{Different Translator Backends}
The translators package supports bountiful backends, including Google, Yandex, Microsoft(Bing), Baidu, Alibaba, Tencent, NetEase(Youdao), Sogou, Deepl, etc\citep{UlionTse2020}. In our experiment, we tested user experiences(0-10) scores from different users after using Bil-DOS with Google, Baidu and Bing backends\footnote{Google and Baidu servers can be available due to the high volume of current user requests.}. The results are shown beneath.

\begin{center}
 \begin{tabular}{||c c||} 
 \hline
 Backend & AvgUsrExp\\ [0.5ex] 
 \hline\hline
 Google & 7.86 \\ 
 \hline
 Baidu & 7.51 \\
 \hline
 Bing & 7.54 \\
 \hline
\end{tabular}
\end{center}

Surprisingly, although there are lags while using Baidu backend, Baidu backend turns out to be almost as good as Bing backend. Baidu translation can be viewed as a translation service specifically designed with Chinese as its source language. So the lag problem can be fixed, Baidu's score would be much higher.

\subsection{Poly-chromatic vs. Mono-chromatic}
Similar to the experiments about translator backends, the comparison of user experiences between different color settings were conducted as well(with Google as translator backend).

\begin{center}
 \begin{tabular}{||c c||} 
 \hline
 Color settings & AvgUsrExp\\ [0.5ex] 
 \hline\hline
 Poly-chromatic & 7.86 \\ 
 \hline
 Mono-chromatic & 7.22 \\
 \hline
\end{tabular}
\end{center}

As can be found from the results, the poly-chromatic UI can boost the user experience scores by almost 10\%.

\subsection{Word-level vs. Phrase-level}
While implementing, we designed two different methods to tokenize utterances and map keywords. The first one is word-level, another one is phrase-level. In word-level, Bil-DOS breaks all user utterances into single words, then maps the keywords from the currently stored resource pool. And in Phrase-level, Bil-DOS keeps the initial user utterance and directly maps the keywords. We defined failure rate as the ratio of the number of task in-completion out of all times of use(3 out of 47 times for word-level, 2 out of 51 times for phrase-level) with a restrained number of turns. In this experiment, we tested Bil-DOS failure rate under these two strategies. Here is the result:

\begin{center}
 \begin{tabular}{||c c||} 
 \hline
 Tokenization\&Mapping & FailureRate\\ [0.5ex] 
 \hline\hline
 Word-level & 6.38\% \\ 
 \hline
 Phrase-level & 3.92\% \\
 \hline
\end{tabular}
\end{center}

The phrase-level tokenization and mapping strategy yields only half of the failure rate compared with the word-level one. However, Bil-DOS needs to solve task in-completion with other more sophisticated strategies.

\begin{figure*}[t]
    \includegraphics[width=0.99\textwidth]{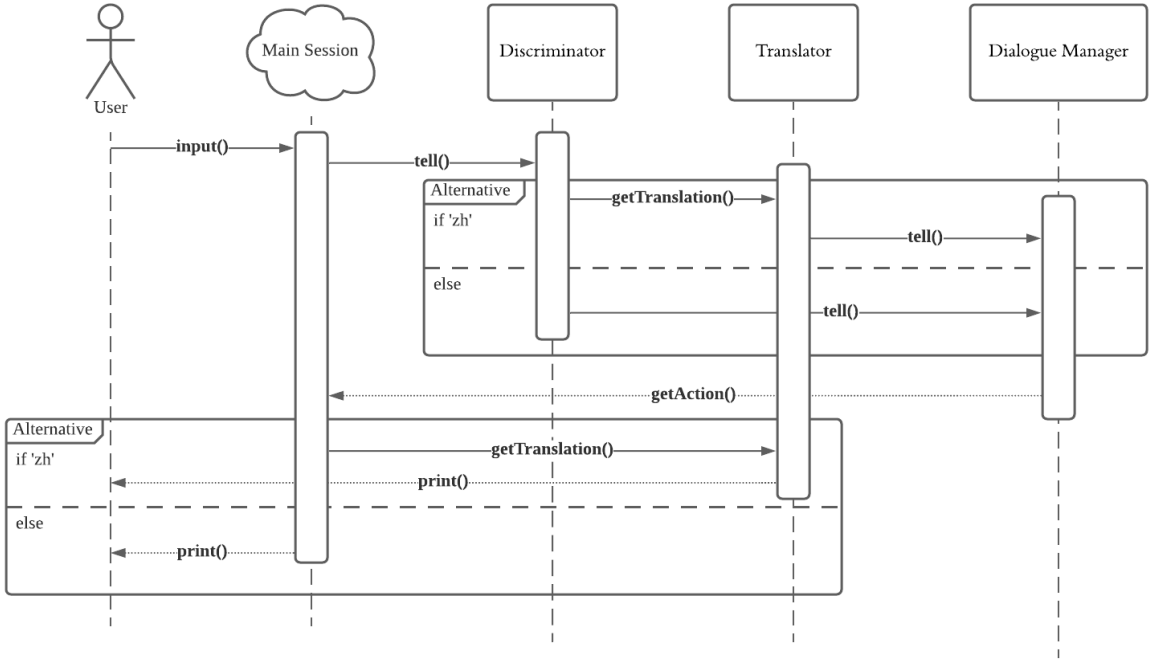}
    \centering
    \caption{UML sequence diagram with main session}
    \label{fig:UMLSeq}
\end{figure*}

\subsection{Robustness and Correctness}
Bil-DOS is designed with an increasing robustness as the times of use increments. We designed another experiment to record the average user experience scores as the times of use increase.

\begin{tikzpicture}
\begin{axis}[
    title={User experience scores over time},
    xlabel={Times of use},
    ylabel={User experience score},
    xmin=0, xmax=10,
    ymin=6, ymax=9,
    xtick={0,2, 4, 6, 8, 10},
    ytick={6, 7, 8, 9},
    legend pos=north west,
    grid style=dashed,
]

\addplot[
    color=red,
    mark=square,
    ]
    coordinates {
    (0,7.12)(1,7.27)(2,7.51)(3,7.69)(4,7.81)(5,7. 86)(6,7.80)(7,7.71)(8,7.75)(9,7.68)(10,7.65)
    };
    \legend{AvgUsrExp}
    
\end{axis}
\end{tikzpicture}

After experiments, the human-in-the-loop annotation did help the conversation process at the beginning, however, as the times of use increase, the robustness does not always mean correctness and user satisfaction. This phenomenon is quite similar to the relation between recall, precision and F-1 score. More robustness here means higher recall, however, higher recall will not bring a higher F-1 score every time.

\section{Error Analysis and Conclusion}
After experiments, although Bil-DOS got fairly good evaluation results, it still was not able to handle more complex situations. In this section, we will discuss some unhappy cases that Bil-DOS is not able to handle in the current version.

\subsection{Imbalanced Attention}
Bil-DOS tends to give more attention to the currently stored intentions and keywords. Because, in the mapping process, Bil-DOS always uses the stored keywords, keeps them fixed, and iterates the inputted sentences to seek the matched patterns. So no matter what kind of tokenization or mapping strategy is being used, the stored keywords will receive more attention than the words in user utterance. One unhappy case we encountered was every-time when the word `\emph{Chinese}' is entered, the keyword detected will be '\emph{hi}' and the matched intention will be `\emph{greet}'. We think a good solution to it might be introducing some statistical learning classification methods like Maximum Likelihood Estimation \citep{mccallum1998improving, xu2003representative}, Naive Bayes \citep{mccallum1998comparison, kim2006some}, Support Vector Machine \citep{sun2009strategies, wang2006optimal} or even Neural Nets \citep{lai2015recurrent, kim2014convolutional}. 

\subsection{Co-reference, Polysemy}
In Linguistics, the term polysemy refers to the phenomenon where one word carries multiple meanings. In Bil-DOS, polysemy means a little bit different. The easiest example, in the menu, the word `\emph{Italian}' can refer to the keyword `\emph{Italian} bread' with the intention bread, or the keyword `\emph{Italian} cheese' with the intention cheese. However, neither word-level nor phrase level strategy is able to handle this situation. In the mapping process in both strategies, Bil-DOS maintains a hash map to track and store relations, which can mean complex situations like co-reference or polysemy will confuse Bil-DOS and lead to task in-completion. 

\subsection{Synonym}
Due to the lack of ground knowledge, every time when Bil-DOS meets a synonym of any stored keyword, it will emit the human-in-the-loop annotation process. Because, Bil-DOS only analyzes morphology instead of semantics of each word. And it might lead to redundant annotation. Solutions to the lack of ground knowledge in Machine Learning can be rule based methods \citep{liu2020lifelong}, pre-trained word embeddings \citep{pennington2014glove, miller1995wordnet} or even applying BERT \citep{devlin2018bert} as prior knowledge. 

\begin{CJK*}{UTF8}{gbsn}
Some translator backends also show no consistency when translating synonyms, which might make the human-in-the-loop annotation process more often to take place. For example, the word `\emph{Avocado}' can be translated to `牛油果(can be translated as `ao fruit' by some translators)' or `鳄梨'. If the user has already re-annotated the one of the word and enters its synonym, due to the translation inconsistency, the user has to go through the re-annotation process again.
\end{CJK*}

\section{Future Work}
We have finished some basic functions in Bil-DOS, and it is capable of conducting conversations with users to help them with their orders. However, Bil-DOS still needs refinements to completely meet user requirements. Here we list several unimplemented ideas to make Bil-DOS better.

\subsection{Break Turn-level Analysis}
As stated before, currently, all the modules in Bil-DOS support only turn-level analysis, which means it only memorizes the information within each turn and throws the stored information once the turn is finished. So it fails to analyze the relations between different turns. More context information between turns should be considered in the further versions to make Bil-DOS better understand what the user is trying to convey. However, when there are more than one turns that need to be analyzed, the current naive keyword-based search might not work well. Thus, a better or more mathematically sophisticated approach, like LSTM \citep{hochreiter1997long} or RCNN \citep{lai2015recurrent}, is needed to analyze the semantics, map with certain keywords, and classify into certain intentions. 

\subsection{All in Two}
All targeted users of Bil-DOS are supposed to speak English AND Chinese. Thus, the human-in-the-loop annotation process is currently not in Chinese, which means the process needs the user to be able to use English. In order to serve as many users as possible, Bil-DOS needs to support two different languages in any process at any time. In future versions, we would like to make the system completely bi-lingual, which means all processes in the system are able to take two different languages. By doing this, users will really be able to arbitrarily switch between languages in any process at any time. However, this idea might make more requirements about the translation process and make the execution sequence even more complex.

\subsection{Plot More Stories}
Although Bil-DOS is able to take all kinds of user utterances with all kinds of keywords and all kinds of intentions, the stored ordering process still remains a fixed story, which is `\emph{bread}', `\emph{cheese}', `\emph{vegetable}', `\emph{sauce}' and `\emph{extra}'. When an irrelevant intention is involved in the user utterance, the intention will be detected and stored locally, and will never be able to participate in the ordering process. So the human-in-the-loop process can be useless for ordering while handling irrelevant intentions. However, we would like Bil-DOS to be capable of handling more stories other than Subway ordering only. In order to do that, Bil-DOS might need another module called {\verb|StroyWeaver|} to customize the stories for all different situations.

\subsection{Teamwork Allocation}
\begin{center}
 \begin{tabular}{||c | c||} 
 \hline
 Job Done & By Whom\\ [0.5ex] 
 \hline\hline
 Preliminary Design & Zirong Chen; Haotian Xue \\ 
 \hline
 Translator & Haotian Xue \\
 \hline
 Discriminator & Zirong Chen \\
 \hline
 Dialogue Manager & Zirong Chen; Haotian Xue \\
 \hline
 Main Session & Zirong Chen; Haotian Xue \\
 \hline
 Experiments & Haotian Xue \\
 \hline
 GitHub Maintenance & Zirong Chen \\
 \hline
\end{tabular}
\end{center}

\section*{Acknowledgments}
Our work is designed for the final project for the course Dialogue System(COSC 483, Fall 2020) at Georgetown University. We would especially thank Ziyao Ding from Georgetown University, Yuansheng Xie from Dartmouth College, Hanyue Gu from UIUC for their participation in the experiments and the improvements for Bil-DOS. We would also like to thank Professor Matthew Marge for his support and assistance.

\bibliography{acl2020}
\bibliographystyle{acl_natbib}

\end{document}